%% file: main.tex
\documentclass[11pt,a4paper]{article}
\usepackage[hyperref]{emnlp-ijcnlp-2019}
\usepackage{times}
\usepackage{latexsym}
\usepackage{url}
\usepackage{enumitem}
\usepackage{stfloats}
\usepackage{float}
\usepackage{nccmath}
\usepackage{mathtools}
\usepackage{epstopdf}
\usepackage{amsfonts}
\usepackage{amssymb}
\usepackage{CJKutf8}
\usepackage{CJK}
\usepackage{amsmath}
\usepackage{verbatim}
\usepackage{multirow}
\usepackage{subfigure}
\usepackage{pifont}
\usepackage{array}
\usepackage{fp}
\usepackage{makecell}
\usepackage{flushend}
\usepackage{cases}
\usepackage{makecell}
\usepackage{color, soul}
\usepackage{xcolor}
\usepackage{bm}
\usepackage{booktabs}
\usepackage{tablefootnote}
\usepackage{tcolorbox}
\usepackage{siunitx}
\usepackage[linesnumbered,ruled,vlined]{algorithm2e}
\usepackage[noend]{algpseudocode}
\usepackage{graphicx}
\aclfinalcopy 

\title{Read, Attend and Comment: A Deep Architecture for Automatic News Comment Generation}

\author{
  Ze Yang$^\dag$, Can Xu$^\diamondsuit$, Wei Wu$^\diamondsuit$, Zhoujun Li$^\dag$\thanks{~~~Corresponding Author}~~~~~\\
  $^\dag$State Key Lab of Software Development Environment, Beihang University, Beijing, China\\
  $^\diamondsuit$Microsoft Corporation, Beijing, China\\
  \{tobey, lizj\}@buaa.edu.cn 
  \{wuwei, caxu\}@microsoft.com
}

\date{}

\newtcbox{\mybox}[1][red]
  {on line, arc = 0pt, outer arc = 0pt,
    colback = #1!10!white, colframe = #1!50!black,
    boxsep = 0pt, left = 0pt, right = 0pt, top = 1.8pt, bottom = 1.8pt,
    boxrule = 0pt, bottomrule = 0.5pt, toprule = 0.5pt}

\begin{document}
\maketitle
\begin{abstract}
  Automatic news comment generation is a new testbed for techniques of natural language generation. In this paper, we propose a ``read-attend-comment'' procedure for news comment generation and formalize the procedure with a reading network and a generation network. The reading network comprehends a news article and distills some important points from it, then the generation network creates a comment by attending to the extracted discrete points and the news title. We optimize the model in an end-to-end manner by maximizing a variational lower bound of the true objective using the back-propagation algorithm. Experimental results on two datasets indicate that our model can significantly outperform existing methods in terms of both automatic evaluation and human judgment.
\end{abstract}

\input{introduction.tex}   
\input{related_work.tex}
\input{approach.tex}

\input{experiment.tex}

\input{conclusion.tex}

\bibliography{emnlp-ijcnlp-2019}
\bibliographystyle{acl_natbib}

\input{appendix.tex}

\end{document}

%% file: introduction.tex
    \section{Introduction}
	

In this work, we study the problem of automatic news comment generation, which is a less explored task in the literature of natural language generation  \cite{gatt2018survey}. We are aware that numerous uses of these techniques can pose ethical issues and that best practices will be necessary for guiding applications. In particular, we note that people expect comments on news to be made by people. Thus, there is a risk that people and organizations could use these techniques at scale to feign comments coming from people for purposes of political manipulation or persuasion.  In our intended target use, we explicitly disclose the generated comments on news as being formulated automatically by an entertaining and engaging chatbot \cite{shum2018eliza}. Also, we understand that the behaviors of deployed systems may need to be monitored and guided with methods, including post-processing techniques \cite{van2018challenges}. While there are risks with this kind of AI research, we believe that developing and demonstrating such techniques is important for understanding valuable and potentially troubling applications of the technology.
	
	\begin{table}[!t]
	\small
    \centering
    \begin{tabular}{p{7.3cm}}
    \toprule
    \textbf{Title:}\space FIFA rankings: France number one, Croatia and England soar, Germany and Argentina plummet\\
    \midrule
    \textbf{Body (truncated):}\space World Cup glory has propelled France to the top of FIFA's latest world rankings, with the impact of Russia 2018 felt for better or worse among a number of football's heavyweight nations.\\
     These are the first set of rankings released under FIFA's new formula that "relies on adding/subtracting points won or lost for a game to/from the previous point totals rather than averaging game points over a given time period".\\
    FIFA world rankings:
    1.France \space 2.Belgium \space 3. Brazil \space 4. Croatia \space 5. Uruguay \space 6. England \space 7. Portugal \space 8. Switzerland \space 9. Spain \space 10. Denmark\\
    \midrule
    \textbf{Comment A:}\space If it's heavily based on the 2018 WC, hence England leaping up the rankings, how are Brazil at 3?\\
    \textbf{Comment B:}\space England above Spain, Portugal and Germany. Interesting.\\
    \bottomrule 
    \end{tabular}
    \caption{A news example from Yahoo!}
    \label{tab:news_example}
    \end{table}

	Existing work on news comment generation includes preliminary studies, where a comment is generated either from the title of a news article only \cite{zheng2018automatic,qin2018automatic} or by feeding the entire article (title plus body) to a basic sequence-to-sequence (s2s) model with an attention mechanism \cite{qin2018automatic}. News titles are short and succinct, and thus only using news titles may lose quite a lot of useful information in comment generation. On the other hand,  a news article and a comment is not a pair of parallel text. The news article is much longer than the comment and contains much information that is irrelevant to the comment. Thus, directly applying the s2s model, which has proven effective in machine translation \cite{bahdanau2014neural}, to the task of news comment generation is unsuitable, and may bring a lot of noise to generation. Both approaches oversimplify the problem of news comment generation and are far from how people behave on news websites. In practice, people read a news article, draw attention to some points in the article, and then present their comments along with the points they are interested in. Table \ref{tab:news_example} illustrates news commenting with an example from Yahoo! News.\footnote{\url{https://www.yahoo.com/news/fifa-rankings-france-number-one-112047790.html}} The article is about the new FIFA ranking, and we pick two comments among the many to explain how people behave in the comment section. First, both commenters have gone through the entire article, as their comments are built upon the details in the body. Second, the article gives many details about the new ranking, but both commenters only comment on a few points. Third, the two commenters pay their attention to different places of the article: the first one notices that the ranking is based on the result of the new world cup, and feels curious about the position of Brazil; while the second one just feels excited about the new position of England. The example indicates a ``read-attend-comment'' behavior of humans and sheds light on how to construct a model.
	
    We propose a reading network and a generation network that generate a comment from the entire news article. The reading network simulates how people digest a news article, and acts as an encoder of the article. 
	The generation network then simulates how people comment the article after reading it, and acts as a decoder of the comment.
	Specifically, from the bottom to the top, the reading network consists of a representation layer, a fusion layer, and a prediction layer. The first layer represents the title of the news with a recurrent neural network with gated recurrent units (RNN-GRUs) \cite{cho2014properties} and represents the body of the news through self-attention which can model long-term dependency among words. The second layer forms a representation of the entire news article by fusing the information of the title into the representation of the body with an attention mechanism and a gate mechanism. The attention mechanism selects useful information in the title, and the gate mechanism further controls how much such information flows into the representation of the article. Finally, the third layer is built on top of the previous two layers and employs a multi-label classifier and a pointer network \cite{vinyals2015pointer} to predict a bunch of salient spans (e.g., words, phrases, and sentences, etc.) from the article. With the reading network, our model comprehends the news article and boils it down to some key points (i.e., the salient spans).  The generation network is an RNN language model that generates a comment word by word through an attention mechanism \cite{bahdanau2014neural} on the selected spans and the news title. In training, since salient spans are not explicitly available, we treat them as a latent variable, and jointly learn the two networks from article-comment pairs by optimizing a lower bound of the true objective through a Monte Carlo sampling method. Thus, training errors in comment prediction can be back-propagated to span selection and used to supervise news reading comprehension.
	
	We conduct experiments on two large scale datasets. One is a Chinese dataset published recently in \cite{qin2018automatic}, and the other one is an English dataset built by crawling news articles and comments from Yahoo! News. Evaluation results on the two datasets indicate that our model can significantly outperform existing methods on both automatic metrics and human judgment.
	
	Our contributions are three-folds: (1) proposal of ``read-attend-comment" procedure for news comment generation with a reading network and a generation network; (2) joint optimization of the two networks with an end-to-end learning approach; and (3) empirical verification of the effectiveness of the proposed model on two datasets. 

%% file: related_work.tex
\section{Related Work}
News comment generation is a sub-task of natural language generation (NLG). Among various NLG tasks, the task studied in this paper is most related to summarization \cite{rush2015neural,nallapati2016abstractive,see2017get} and product review generation \cite{tang2016context,dong2017learning}. However, there is stark difference between news comment generation and the other two tasks: the input of our task is an unstructured document, while the input of product review generation is structured attributes of a product; and the output of our task is a comment which often extends the content of the input with additional information, while the output of summarization is a condensed version of the input that contains the main information from the original. Very recently, there emerge some studies on news comment generation. For example, \newcite{zheng2018automatic} propose a gated attention neural network model to generate news comments from news titles. The model is further improved by a generative adversarial net. \newcite{qin2018automatic} publish a dataset with results of some basic models. Different from all the existing methods, we attempt to comprehend the entire news articles before generation and perform end-to-end learning that can jointly optimize the comprehension model and the generation model.

Our model is partially inspired by the recent success of machine reading comprehension (MRC), whose prosperity can be attributed to an increase of publicly available large scale annotated datasets, such as SQuAD \cite{rajpurkar2016squad,rajpurkar2018know} and MS Marco \cite{nguyen2016ms} etc. A great number of models have been proposed to tackle the MRC challenges, including BiDAF \cite{seo2016bidirectional}, r-net \cite{wang2017gated}, DCN \cite{xiong2016dynamic}, Document Reader \cite{chen2017reading}, QANet \cite{yu2018qanet}, and s-net \cite{tan2017s} etc. Our work can be viewed as an application of MRC to a new NLG task. The task aims to generate a comment for a news article, which is different from existing MRC tasks whose goal is to answer a question. Our learning method is also different from those in the MRC works. 

%% file: approach.tex
\section{Approach}
\subsection{Problem Formalization}

Suppose we have a dataset $\mathcal{D}=\{(T_i,B_i,C_i)\}^N_{i=1}$, where the $i$-th triple $(T_i,B_i,C_i)$ consists of a news title $T_i$, a news body $B_i$, and a comment $C_i$. Our goal is to estimate a probability distribution $P(C| T, B)$ from $\mathcal{D}$, and thus, given a new article $(T,B)$ with $T$ the news title and $B$ the news body, we can generate a comment $C$ following $P(C| T, B)$.

\begin{figure*}[t!]
    \centering
    \includegraphics[width=4.8in]{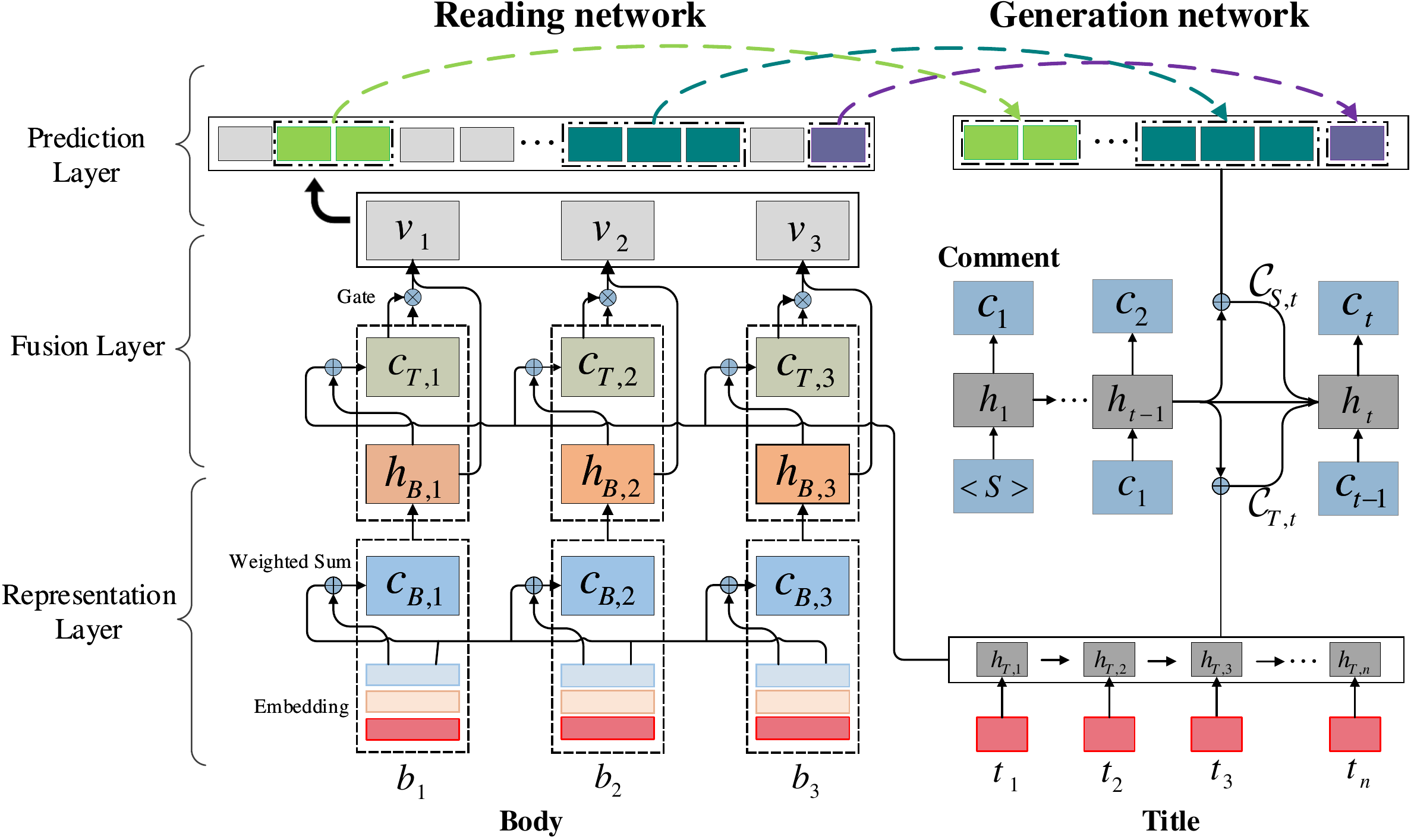}
    \caption{Architecture of our model. The black solid arrows represent differentiable operations and the dashed arrows are non-differentiable operations which represent distilling points from news body.}
    \label{fig:model}
\end{figure*}

\subsection{Model Overview}
Figure \ref{fig:model} illustrates the architecture of our model. In a nutshell, the model consists of a reading network and a generation network. The reading network first represents a news title and a news body separately in a representation layer, then forms a representation of the entire article by fusing the title into the body through a fusion layer, and finally distills some salient spans from the article by a prediction layer. The salient spans and the news title are then fed to the generation network to synthesize a comment. With the two networks, we can factorize the generation probability $P(C| T, B)$ as $P(S| T, B) \cdot P(C| S, T)$, where $S=(s_1,\ldots, s_w)$ refers to a set of spans in $B$, $P(S| T, B)$ represents the reading network, and $P(C| S, T)$ refers to the generation network.


\subsection{Reading Network}
In the representation layer, let $T=(t_1,\ldots,t_n)$ be a news title with $t_j$ the $j$-th word, and $B=(b_1,\ldots,b_m)$ be the associated news body with $b_k$ the $k$-th word,  we first look up an embedding table and represent $t_j$ and $b_k$ as $e_{T,j} \in \mathbb{R}^{d_1}$ and $e_{B,k} \in \mathbb{R}^{d_1}$ respectively, where $e_{T,j}$ and $e_{B,k}$ are randomly initialized, and jointly learned with other parameters. Different from the title, the body is long and consists of multiple sentences. Hence, to emphasize positional information of words in the body, we further expand $e_{B,k}$ with $o_{B,k}$ and $s_{B,k}$, where $o_{B,k}, s_{B,k} \in \mathbb{R}^{d_2}$ are positional embeddings with the former indexing the position of $b_k$ in its sentence and the latter indicating the position of the sentence in the entire body. The representation of $b_k$ is then given by $\hat{e}_{B,k}=\text{MLP}([e_{B,k};o_{B,k}; s_{B,k}])$, where $\hat{e}_{B,k}\in \mathbb{R}^{d_1}$, $\text{MLP}(\cdot)$ refers to a multi-layer perceptron with two layers, and $[\cdot;\cdot;\cdot]$ means the concatenation of the three arguments. 

Starting from $\mathcal{E}_T=(e_{T,1}, \ldots, e_{T,n})$ and $\mathcal{E}_B=(\hat{e}_{B,1},\ldots,\hat{e}_{B,m})$ as initial representations of $T$ and $B$ respectively, the reading network then transforms $T$ into a sequence of hidden vectors $\mathcal{H}_T=(h_{T,1},\ldots, h_{T,n})$ with a recurrent neural network with gated recurrent units (RNN-GRUs) \cite{cho2014properties}. 
In the meanwhile, $B$ is transformed to $\mathcal{H}_B=(h_{B,1},\ldots, h_{B,m})$ with the $k$-th entry $h_{B,k} \in \mathbb{R}^{d_1}$ defined as $\text{MLP}([\hat{e}_{B,k}; c_{B,k}])$ (two layers) and  $c_{B,k}$ is an attention-pooling vector calculated by a scaled dot-product attention \cite{vaswani2017attention} denoted as $\textit{dot-att}(\mathcal{E}_B,\hat{e}_{B,k})$:
\begin{equation}
\small
\label{attend-pool}
    \begin{split}
        &\textstyle{c_{B,k}=\sum^m_{j=1}\alpha_{k,j} \hat{e}_{B,j}},\\ &\textstyle{\alpha_{k,j}=\text{exp}(s_{k,j})/\sum^m_{l=1}\text{exp}(s_{k,l})},\\
        &s_{k,j}=({\hat{e}_{B,k}}^\top \hat{e}_{B,j})/\sqrt{d_1}.
    \end{split}
\end{equation}
In Equation (\ref{attend-pool}), a word is represented via all words in the body weighted by their similarity. By this means, we try to capture the dependency among words in long distance. 

The fusion layer takes $\mathcal{H}_T$ and $\mathcal{H}_B$ as inputs, and produces $\mathcal{V}=(v_1,\ldots,v_m)$ as new representation of the entire news article by fusing $\mathcal{H}_T$ into $\mathcal{H}_B$. Specifically, $\forall h_{B,k} \in \mathcal{H}_B$, we first let  $h_{B,k}$ attend to $\mathcal{H}_T$ and form a representation $c_{T,k}=\textit{dot-att}(\mathcal{H}_T, h_{B,k})$, where $\textit{dot-att}(\mathcal{H}_T, h_{B,k})$ is parameterized as Equation (\ref{attend-pool}). With this step, we aim to recognize useful information in the title. Then we combine $c_{T,k}$ and $h_{B,k}$ as $v_k \in \mathbb{R}^{d_1}$ with a gate $g_k$ which balances the impact of $c_{T,k}$ and $h_{B,k}$ and filters noise from $c_{T,k}$. $v_k$ is defined by:
\begin{equation}
\small
    \begin{split}
        &v_k=h_{B,k}+ g_k\odot c_{T,k},\\
        &g_k=\sigma(W_g[h_{B,k};c_{T,k}]),    
    \end{split}
\end{equation}

The top layer of the reading network extracts a bunch of salient spans based on $\mathcal{V}$. Let $S=((a_1,e_1),\ldots,(a_w,e_w))$ denote the salient spans, where $a_i$ and $e_i$ refer to the start position and the end position of the $i$-th span respectively, we propose detecting the spans with a multi-label classifier and a pointer network. Specifically, we formalize the recognition of $(a_1, \ldots, a_w)$ as a multi-label classification problem with  $\mathcal{V}$ as an input and  $L=(l_1,\ldots,l_m)$ as an output, where $\forall k \in \{1,\ldots,m\}$, $l_k=1$ means that the $k$-th word is a start position of a span, otherwise $l_k=0$. Here, we assume that $(a_1, \ldots, a_w)$ are independent with each other, then the multi-label classifier can be defined as $m$ binary classifiers with the $k$-th classifier given by 
\begin{equation}
\small
\hat{y}_k=\text{softmax}(\text{MLP}_k(v_k)),
\label{multilabel}
\end{equation}
where $\text{MLP}_k(\cdot)$ refers to a two-layer MLP, and $\hat{y}_k\in \mathbb{R}^2$ is a probability distribution with the first entry as $P(l_k=0)$ and the second entry as $P(l_k=1)$. 
The advantage of the approach is that it allows us to efficiently and flexibly detect a variable number of spans from a variable-length news article, as there is no dependency among the $m$ classifiers, and they can be calculated in parallel.

Given $a_k$, the end position $e_k$ is recognized via a probability distribution $(\alpha_{a_k,1},\ldots, \alpha_{a_k,m})$ which is defined by a pointer network:
\begin{equation}
\small
\label{pointer}
    \begin{split}
        &\textstyle{\alpha_{a_k,j}=\text{exp}(s_{a_k,j})/\sum^m_{l=1}\text{exp}(s_{a_k,l})},\\
        &s_{a_k,j}=V^\top \text{tanh}(W_v v_j+W_h h_{a_k,1}),\\
        &h_{a_k,1}=\text{GRU}(h_0,[c_0; v_{a_k}]),\\
    \end{split}
\end{equation}
where $h_0=att(\mathcal{V},r)$ is an attention-pooling vector based on parameter $r$:
\begin{equation}
\small
\label{att}
    \begin{split}
        &\textstyle{h_0=\sum_{j=1}^m \beta_j \cdot v_j},\\
        &\textstyle{\beta_j=\text{exp}(\beta'_j)/\sum^m_{l=1}\text{exp}(\beta'_l)},\\
        &\beta'_j=V_1^\top \text{tanh}(W_{v,1} v_j+W_{h,1} r),\\
    \end{split}
\end{equation}
$c_0=att(\mathcal{V},h_0)$ is defined in a similar way with $r$ replaced by $h_0$.

Let us denote $(a_1,\ldots, a_w)$ as $start$ and $P(l_i=1)$ as $p_i$, then $P(S|T,B)$ can be formulated as
\begin{equation}
\small
\label{p_s}
    P(S|T,B) = \prod_{k=1}^w [p_{a_k} \cdot \alpha_{a_k, e_k}] \prod_{i\notin start, 1\le i \le m} [1-p_i].
\end{equation}
In practice, we recognize the $i$-th word as a start position if $p_i>1-p_i$, and determine the associated end position by $\arg\max_{1\le k \le m} \alpha_{i,k}$. Note that we do not adopt a pointer network to detect the start positions, because in this case, either we have to set a threshold on the probability distribution, which is sensitive to the length of the news article and thus hard to tune, or we can only pick a fixed number of spans for any articles by ranking the probabilities. Neither of them is favorable.

\subsection{Generation Network}
With $S=((a_1,e_1),\ldots,(a_w,e_w))$ the salient spans, $\mathcal{V}=(v_1,\ldots,v_m)$ the representation of the news article, and $\mathcal{H}_T$ the representation of the news title given by the three layers of the reading network respectively, we define a representation of $S$ as $\mathcal{H}_S=(v_{a_1},v_{a_1+1},\ldots, v_{e_1}, \ldots, v_{a_w}, v_{a_w+1},\ldots, v_{e_w})$. The generation network takes $\mathcal{H}_T$ and $\mathcal{H}_S$ as inputs and decodes a comment word by word via attending to both $\mathcal{H}_T$ and $\mathcal{H}_S$. At step $t$, the hidden state is $h_{t}=\text{GRU}(h_{t-1}, [e_{C,{t-1}}; \mathcal{C}_{T,{t-1}}; \mathcal{C}_{S,{t-1}}])$, $h_t \in \mathbb{R}^{d_1}$. $e_{C,{t-1}}$ is the embedding of the word generated at step $t$-$1$, $\mathcal{C}_{T,{t-1}}=att(\mathcal{H}_T, h_{t-1})$ and $\mathcal{C}_{S,{t-1}}=att(\mathcal{H}_S, h_{t-1})$ are context vectors that represent attention on the title and the spans respectively. $att(\cdot,\cdot)$ is defined as Equation (\ref{att}).

With $h_t$, we calculate $\mathcal{C}_{T,t}$ and $\mathcal{C}_{S,t}$ via $att(\mathcal{H}_T, h_{t})$ and $att(\mathcal{H}_S, h_t)$ respectively and obtain a probability distribution over vocabulary by $\mathcal{P}_t=\text{softmax}(V[h_t; \mathcal{C}_{T,t}; \mathcal{C}_{S,t}]+b)$. Let $C=(c_1,\ldots,c_o)$ be a comment where $\forall k\in \{1,\ldots, o\}$, $c_k$ is the index of the $k$-th word of $C$ in vocabulary, then $P(C |S, T)$ is defined as

\begin{equation}\nonumber
\small
\begin{split}
\label{p_c}
    P(C| S, T) &=P(c_1| S, T) \prod^o_{t=2} P(c_t|c_1,\ldots,c_{t-1}, S, T)\\
    &=\mathcal{P}_1(c_1) \prod^o_{t=2} \mathcal{P}_t(c_t),\\
    \end{split}
\end{equation}
where $\mathcal{P}_t(c_t)$ refers to the $c_t$-th entry of $\mathcal{P}_t$. In decoding, we define the initial state $h_{0}$ as an attention-pooling vector over the concatenation of $\mathcal{H}_T$ and $\mathcal{H}_S$ given by $att([\mathcal{H}_T;\mathcal{H}_S], q)$. $q$ is a parameter learned from training data.

\subsection{Learning Method}
We aim to learn $P(S|T,B)$ and $P(C|S,T)$ from $\mathcal{D}=\{(T_i,B_i,C_i)\}^N_{i=1}$, but $S$ is not explicitly available, which is a common case in practice. To address the problem, we treat $S$ as a latent variable, and consider the following objective:
\begin{equation}
\small
\label{obj1}
    \begin{split}
    \mathcal{J} &= \sum_{i=1}^N \log  P(C_i | T_i, B_i)\\
    &=\sum_{i=1}^N \log \Big(\sum_{S_i \in \mathcal{S}} P(S_i | T_i, B_i) P(C_i | S_i, T_i \Big),
    \end{split}
\end{equation}
where $\mathcal{S}$ refers to the space of sets of spans, and $S_i$ is a set of salient spans for $(T_i, B_i)$. Objective $\mathcal{J}$ is difficult to optimize, as logarithm is outside the summation. Hence, we turn to maximizing a lower bound of Objective $\mathcal{J}$ which is defined as:
\begin{equation}
\small
\label{obj2}
    \begin{split}
        \mathcal{L} &= \sum_{i=1}^N \sum_{S_i \in \mathcal{S}} P(S_i| T_i,B_i) \log P(C_i|S_i,T_i) < \mathcal{J}\\
    \end{split}
\end{equation}
Let $\Theta$ denote all parameters of our model and $\frac{\partial \mathcal{L}_i}{\partial \Theta}$ denote the gradient of $\mathcal{L}$ on an example $(T_i,B_i,C_i)$, then $\frac{\partial \mathcal{L}_i}{\partial \Theta}$ is given by
\begin{equation}
\small
\label{grad}
\begin{split}
        \sum_{S_i \in \mathcal{S}} P (S_i |& T_i ,B_i)\Big[\frac{\partial \log P(C_i|S_i,T_i)}{\partial \Theta} +\\  
        &\log P(C_i|S_i,T_i)\frac{\partial \log P(S_i|T_i,B_i)}{\partial \Theta} \Big].
\end{split}
\end{equation}
To calculate the gradient, we have to enumerate all possible $S_i$s for $(T_i,B_i)$, which is intractable. Thus, we employ a Monte Carlo sampling method to approximate $\frac{\partial \mathcal{L}_i}{\partial \Theta}$. Suppose that there are $J$ samples, then the approximation of $\frac{\partial \mathcal{L}_i}{\partial \Theta}$ is given by
\begin{equation}
\small
\label{approx}
    \begin{split}
        \frac{1}{J}\sum_{n=1}^{J}\Big[ & \frac{\partial \log P(C_i|S_{i,n},T_i)}{\partial \Theta} + \\
        &\log P(C_i|S_{i,n},T_i)\frac{\partial \log P(S_{i,n} |T_i,B_i)}{\partial \Theta} \Big],
    \end{split}
\end{equation}
where $\forall n$, $S_{i,n}$ is sampled by first drawing a group of start positions according to Equation (\ref{multilabel}), and then  picking the corresponding end positions by Equations (\ref{pointer})-(\ref{att}). Although the Monte Carlo estimator is unbiased, it typically suffers from high variance. To reduce variance, we subtract baselines from  $\log P(C_i|S_{i,n},T_i)$. Inspired by \cite{mnih2014neural},  we introduce an observation-dependent baseline $\mathcal{B}_\psi(T_i,C_i)$ to capture the systematic difference in news-comment pairs during training. Besides, we also exploit a global baseline $\mathcal{B}$ to further control the variance of the estimator. The approximation of $\frac{\partial \mathcal{L}_i}{\partial \Theta}$ is then re-written as 
\begin{equation}
\small
\label{approx2}
    \begin{split}
          &\frac{1}{J}\sum_{n=1}^{J} \Big[\frac{\partial \log P(C_i|S_{i,n},T_i)}{\partial \Theta} + \Big(\log P(C_i| S_{i,n},T_i)\\ 
         & -\mathcal{B}_\psi(T_i,C_i)-\mathcal{B}\Big) 
        \frac{\partial \log P(S_{i,n} |T_i,B_i)}{\partial \Theta} \Big].
    \end{split}
\end{equation}
To calculate $\mathcal{B}_\psi(T_i,C_i)$, we first encode the word sequences of $T_i$ and $C_i$ with GRUs respectively, and then feed the last hidden states of the GRUs to a three-layer MLP.  $\mathcal{B}$ is calculated as an average of $P(C_i| S_{i,n},T_i)- \mathcal{B}_\psi(T_i,C_i)$ over the current mini-batch. The parameters of the GRUs and the MLP are estimated via $\min_{(T_i,B_i)\in \text{mini-batch}, 1\le n \le J} [\log P(C_i|S_{i,n},T_i)-\mathcal{B}_\psi(T_i,C_i)-\mathcal{B}]^2$.

The learning algorithm is summarized in Algorithm \ref{alg}. To speed up convergence, we initialize our model through pre-training the reading network and the generation network. Specifically, $\forall (T_i, B_i, C_i) \in \mathcal{D}$, we construct an artificial span set $\tilde{S}_i$, and learn the parameters of the two networks by maximizing the following objective:
\begin{equation}
\small
    \label{obj3}
        \sum_{i=1}^N \log  P(\tilde{S}_i | T_i, B_i) + \log P (C_i | T_i, \tilde{S}_i)
\end{equation}
$\tilde{S}_i$ is established in two steps: first, we collect all associated comments for $(T_i, B_i)$, extract n-grams ($1\le \text{n} \le 6$) from the comments, and recognize an n-gram in $B_i$ as a salient span if it exactly matches with one of the n-grams of the comments. Second, we break $B_i$ as sentences and calculate a matching score for a sentence and an associated comment. Each sentence corresponds to a group of matching scores, and if any one of them exceeds $0.4$, we recognize the sentence as a salient span. The matching model is pre-trained with $\{T_i, C_i\}_{i=1}^N$ with $C_i$ as a positive example and a randomly sampled comment from other news as a negative example. In the model, $T_i$ and $C_i$ are first processed by GRUs separately, and then the last hidden states of the GRUs are fed to a three-layer MLP to calculate a score.   
\begin{algorithm}[h!]
    \caption{Optimization Algorithm}
			\SetKwData{Index}{Index}
			\SetKwInput{kwInit}{Init}
			\SetKwInput{kwOutput}{Output}
			\SetKwInput{kwInput}{Input}
			\label{alg}
    \small
    {
		\kwInput{training data $\mathcal{D}$, initial learning rate $lr$, MaxStep, sample number $n$.\\}
		\kwInit{ $\Theta$}Construct $\{\tilde{S}_i\}_{i=1}^N$ and pre-train the model by maximizing Objective (\ref{obj3}).\\
        \Indp
        \Indm
        \While{step $<$ MaxStep}
        {
            Randomly sample a mini-batch $k$ from $\mathcal{D}$.\\
            Compute distributions of start positions.\\
            \For{n $<$ J}
            {
                Sample start positions.\\
                Compute distributions of end positions.\\
                Sample end positions.\\
                Compute the terms related to $S_{i,n}$ in Eq. (\ref{approx2}).
            }
            Compute $\mathcal{B}_\psi(C_i, T_i)$ and $\mathcal{B}$ and finish Eq. (\ref{approx2}).\\
            Update the parameters of the model and $\mathcal{B}_\psi(C_i, T_i)$ with SGD.
           
        }
        \kwOutput{  $\Theta$}
    }
\end{algorithm}

%% file: experiment.tex
\section{Experiments}
We test our model on two large-scale news commenting datasets.
\subsection{Experimental Setup}
The first dataset is a Chinese dataset built from Tencent News (news.qq.com) and published recently in \cite{qin2018automatic}. Each data point contains a news article which is made up of a title and a body, a group of comments, and some side information including upvotes and categories. Each test comment is labeled by two annotators according to a 5-scale labeling criteria presented in Table \ref{tab:judge-score}. All text in the data is tokenized by a Chinese word segmenter Jieba (\url{https://github.com/fxsjy/jieba}). The average lengths of news titles, news bodies, and comments are $15$ words, $554$ words and $17$ words respectively. In addition to the Chinese data, we also build another dataset by crawling news articles and the associated comments from Yahoo! News. Besides upvotes and categories, side information in Yahoo data also includes paragraph marks, WIKI-entities, downvotes, abusevotes, and sentiment tagged by Yahoo!. Text in the data is tokenized by Stanford CoreNLP pipline \cite{manning-EtAl:2014:P14-5}. As pre-processing, we filter out news articles shorter than $30$ words in the body and comments shorter than $10$ words or longer than $100$ words. Then, we remove news articles with less than $5$ comments. If the number of comments of an article exceeds $30$, we only keep top $30$ comments with the most upvotes. On average, news titles, news bodies, and comments contain $12$ words, $578$ words and $32$ words respectively. More information about Yahoo data can be found in Appendix \ref{apd:data}. After the pre-processing, we randomly sample a training set, a validation set, and a test set from the remaining data, and make sure that there is no overlap among the three sets. Table \ref{tab:dataset-statistics} summarizes the statistics of the two datasets. Note that we only utilize news titles, news bodies and comments to learn a generation model in this work, but both datasets allow modeling news comment generation with side information, which could be our future work. 

\begin{table}[t!]
    \centering
    \resizebox{1\linewidth}{!}{
    \begin{tabular}{c r c c c c}
    \toprule
                                   &             & Train    & Dev   & Test  \\
         \midrule
         \multirow{2}{*}{Tencent}  &      \# News & 191,502  & 5,000 & 1,610 \\
                                   &  Avg. \# Cmts per News & 27       & 27    & 27    \\
         \midrule
         \multirow{2}{*}{Yahoo}  &          \# News & 152,355  & 5,000 & 3,160 \\
                                 &     Avg. \# Cmts per News & 20.6     & 20.5  & 20.5  \\
         \bottomrule
    \end{tabular}
    }
    \caption{Statistics of the two datasets.  }
    \label{tab:dataset-statistics}
\end{table}

Following \cite{qin2018automatic}, we evaluate the performance of different models with both automatic metrics and human judgment. In terms of automatic evaluation, we employ BLEU \cite {papineni2002bleu}, METEOR \cite{banerjee2005meteor}, ROUGE \cite{lin2004rouge}, and CIDEr \cite{vedantam2015cider} as metrics on both data. Besides these metrics, \newcite{qin2018automatic} propose human score weighted metrics including W-BLEU, W-METEOR, W-ROUGE and W-CIDEr. These metrics, however, requires human judgment on each comment in the test set. Thus, we only involve results w.r.t. these metrics in Tencent data. As \newcite{qin2018automatic} do not publish their code for metric calculation, we employ a popular NLG evaluation project available at \url{https://github.com/Maluuba/nlg-eval}, and modify the scripts with the scores provided in the data according to the formulas in \cite{qin2018automatic} to calculate all the metrics. In human evaluation, for each dataset, we randomly sample $500$ articles from the test data and recruit three native speakers to judge the quality of the comments given by different models. For every article, comments from all models are pooled, randomly shuffled, and presented to the annotators. Each comment is judged by the three annotators under the criteria in Table \ref{tab:judge-score}. 

\begin{table}[t!]
\small
\resizebox{\linewidth}{!}{
\centering
\begin{tabular}{c p{6.5cm}}
\toprule
\textbf{Score} & \textbf{Criteria} \\
\midrule
5 & Rich in content; attractive; deep insights; new yet relevant viewpoints\\
4 & Highly relevant with meaningful ideas \\
3 & Less relevant; applied to other articles \\ 
2 & Fluent/grammatical; irrelevant \\
1 & Hard to read; Broken language; Only emoji\\
\bottomrule
\end{tabular}
}
\caption{Human judgment criteria}
\label{tab:judge-score}
\end{table}

\subsection{Baselines}
The following models are selected as baselines:

\textbf{Basic models}: the retrieval models and the generation models used in \cite{qin2018automatic} including (1) IR-T and IR-TC: both models retrieve a set of candidate articles with associated comments by cosine of TF-IDF vectors. Then the comments are ranked by a convolutional neural network (CNN) and the top position is returned. The difference is that IR-T only utilizes titles, while IR-TC leverages both titles and news bodies; (2) Seq2seq: the basic sequence-to-sequence model \cite{sutskever2014sequence} that generates a comment from a  title; and (3) Att and Att-TC: sequence-to-sequence with attention \cite{bahdanau2014neural} in which the input is either a  title (Att) or a concatenation of a title and a body (Att-TC). In Seq2seq, Att, and Att-TC, top 1 comment from beam search (beam size=5) is returned. 

\textbf{GANN}: the gated attention neural network proposed in \cite{zheng2018automatic}. The model is further improved by a generative adversarial net. 

We denote our model as ``DeepCom'' standing for ``deep commenter'', as it is featured by a deep reading-commenting architecture. All baselines are implemented according to the details in the related papers and tuned on the validation sets.

\subsection{Implementation Details}
For each dataset,  we form a vocabulary with the top $30$k frequent words in the entire data. We pad or truncate news titles, news bodies, and comments to make them in lengths of $30$, $600$, and $50$ respectively. The dimension of word embedding and the size of hidden states of GRU in all models are set as $256$. In our model, we set $d_1$ as $256$ and $d_2$ (i.e., dimension of the position embedding in the reading network) as $128$. The size of hidden layers in all MLPs is $512$. The number of samples in Monte Carlo sampling is $1$. In pre-training, we initialize our model with a Gaussian distribution $\mathcal{N}(0,0.01)$ and optimize Objective (\ref{obj3}) using AdaGrad \cite{duchi2011adaptive} with an initial learning rate $0.15$ and an initial accumulator value $0.1$. Then, we optimize $\mathcal{L}$ using stochastic gradient descent with a learning rate $0.01$. In decoding, top $1$ comment from beam search with a size of $5$ is selected for evaluation. In IR-T and IR-TC, we use three types of filters with window sizes $1$, $3$, and $5$ in the CNN based matching model. The number of each type of filters is $128$.

\subsection{Evaluation Results}
\begin{table*}[t!]
	\centering
	\resizebox{\linewidth}{!}{%
    \begin{tabular}{c r c c c c c c c c c c c c}
    \toprule
    Dataset                  &  Models  & METEOR & W-METEOR & Rouge\_L & W-Rouge\_L & CIDEr & W-CIDEr & BLEU-1 & W-BLEU-1  & Human & Kappa\\ 
    \midrule
	\multirow{7}{*}{Tencent} &  IR-T     & 0.107 & 0.086 & 0.254 & 0.217 & 0.018 & 0.014 & 0.495 & 0.470 &  2.43 & 0.64\\
                             &  IR-TC    & 0.127 & 0.101 & 0.266 & 0.225 & 0.056 & 0.044 & 0.474 & 0.436 &  2.57 & 0.71\\
                             &  Seq2seq  & 0.064 & 0.047 & 0.196 & 0.150 & 0.011 & 0.008 & 0.374 & 0.320 &  1.68 & 0.83\\
                             &  Att      & 0.080 & 0.058 & 0.246 & 0.186 & 0.010 & 0.007 & 0.481 & 0.453 &  1.81 & 0.79\\
                             &  Att-TC   & 0.114 & 0.082 & 0.299 & 0.223 & 0.023 & 0.017 & 0.602 & 0.551 &  2.26 & 0.69\\
                             &  GANN     & 0.097 & 0.075 & 0.282 & 0.222 & 0.010 & 0.008 & 0.312 & 0.278 &  2.06 & 0.73\\
                             &  \bf DeepCom  &\bf 0.181 & \bf 0.138 & \bf 0.317 & \bf 0.250 &  0.029 & 0.023 & \bf 0.721 & \bf 0.656 &  \bf 3.58 & 0.65\\
    \midrule
    \multirow{7}{*}{Yahoo}   &  IR-T     & 0.114 &  -    & 0.214 & -     & 0.014 & -     & 0.472 & -    &  2.71 & 0.67\\
                             &  IR-TC    & 0.117 &  -    & 0.219 & -     & 0.017 & -     & 0.483 & -    &  2.86 & 0.61\\
                             &  Seq2seq  & 0.061 &  -    & 0.203 & -     & 0.011 & -     & 0.365 & -    &  2.26 & 0.68\\
                             &  Att      & 0.075 &  -    & 0.217 & -     & 0.017 & -     & 0.462 & -    &  2.29 & 0.78\\
                             &  Att-TC   & 0.089 &  -    & 0.246 & -     & 0.022 & -     & 0.515 & -    &  2.74 & 0.63\\
                             &  GANN     & 0.079 &  -    & 0.228 & -     & 0.019 & -     & 0.496 & -    &  2.52 & 0.64\\
                             &  \bf DeepCom  & 0.107 &  - & \bf 0.263 &  - & \bf 0.024 &  -  & \bf 0.665 &  - & \bf 3.35 & 0.68\\   
    \bottomrule
	\end{tabular}
	}
	\caption{Evaluation results on automatic metrics and human judgment. Human evaluation results are calculated by combining labels from the three judges. ``Kappa'' means Fleiss' kappa. Numbers in bold mean that improvement over the best baseline is statistically significant.}
    \label{tab:auto-metrics}
\end{table*}

\begin{table}[t!]
\centering
\resizebox{1\linewidth}{!}{
    \centering	
    \begin{tabular}{c r c c c c}
        \toprule
        Dataset &   Metrics  & No Reading & No Prediction & No Sampling & Full Model\\ 
        \midrule
        \multirow{8}{*}{Tencent}& METEOR   & 0.096 & 0.171 & 0.171 & 0.181\\
                                & W-METEOR & 0.072 & 0.129 & 0.131 & 0.138\\
                                & Rouge\_L & 0.282 & 0.307 & 0.303 & 0.317\\
                                & W-Rouge\_L & 0.219 & 0.241 & 0.239 & 0.250\\
                                & CIDEr    & 0.012 & 0.024 & 0.026 & 0.029\\
                                & W-CIDEr  & 0.009 & 0.019 & 0.021 & 0.023\\
                                & BLEU-1   & 0.426 & 0.674 & 0.667 & 0.721\\
                                & W-BLEU-1 & 0.388 & 0.614 & 0.607 & 0.656\\
        \midrule
        \multirow{4}{*}{Yahoo}  & METEOR   & 0.081 & 0.092 & 0.102 & 0.107\\
                                & Rouge\_L & 0.232 & 0.245 & 0.244 & 0.263\\
                                & CIDEr    & 0.017 & 0.023 & 0.020 & 0.024\\
                                & BLEU-1   & 0.490 & 0.531 & 0.609 & 0.665\\
    \bottomrule
    \end{tabular}
    }
    \caption{Model ablation results}		
    \label{tab:ablation}
\end{table}

Table \ref{tab:auto-metrics} reports evaluation results in terms of both automatic metrics and human annotations. On most automatic metrics, DeepCom outperforms baseline methods, and the improvement is statistically significant (t-test with $p$-value $<0.01$). The improvement on BLEU-1 and W-BLEU-1 is much bigger than that on other metrics. This is because BLEU-1 only measures the proportion of matched unigrams out of the total number of unigrams in the generated comments. In human evaluation, although the absolute numbers are different from those reported in \cite{qin2018automatic} due to the difference between human judgements, the overall trend is consistent. In human evaluation, the values of Fleiss' kappa over all models are more than $0.6$, indicating substantial agreement among the annotators. Although built in a complicated structure, GANN does not bring much improvement over other baseline methods, which demonstrates that only using news titles is not enough in comment generation. IR-TC and Att-TC represent the best retrieval model and the best generation model among the baselines on both datasets, implying that news bodies, even used in a simple way, can provide useful information to comment generation.  

\subsection{Discussions}
\paragraph{Ablation study:} We compare the full model of DeepCom with the following variants: (1) No Reading: the entire reading network is replaced by a TF-IDF based keyword extractor, and top $40$ keywords (tuned on validation sets) are fed to the generation network; (2) No Prediction: the prediction layer of the reading network is removed, and thus the entire $\mathcal{V}$ is used in the generation network; and (3) No Sampling: we directly use the model pre-trained by maximizing Objective (\ref{obj3}). Table \ref{tab:ablation} reports the results on automatic metrics. We can see that all variants suffer from performance drop and No Reading is the worst among the three variants. Thus, 
we can conclude that (1) span prediction cannot be simply replaced by TF-IDF based keyword extraction, as the former is based on a deep comprehension of news articles and calibrated in the end-to-end learning process; (2) even with sophisticated representations, one cannot directly feed the entire article to the generation network, as comment generation is vulnerable to the noise in the article; and (3) pre-training is useful, but optimizing the lower bound of the true objective is still beneficial. 

To further understand why DeepCom is superior to its variants, we calculate BLEU-1 (denoted as BLEU$_{\text{span}}$) with the predicted spans and the ground truth comments in the test sets of the two data, and compare it with a baseline BLEU-1 (denoted as BLEU$_{\text{base}}$) which is calculated with the entire news articles and the ground truth comments. On Tencent data, BLEU$_{\text{span}}$ and BLEU$_{\text{base}}$ are $0.31$ and $0.17$ respectively, and the two numbers on Yahoo data are $0.29$ and $0.16$ respectively. This means that by extracting salient spans from news articles, we can filter out redundant information while keeping the points that people like to comment on, which explains why DeepCom is better than No Prediction. When comparing DeepCome with No Sampling, we find that spans in DeepCom are longer than those in No Sampling. In the test set of Tencent data, the average lengths of salient spans with and without sampling are $11.6$ and $2.6$ respectively, and the two numbers in Yahoo data are $14.7$ and $2.3$ respectively. Thus, DeepCom can leverage discourse-level information rather than a few single words or bi-grams in comment generation, which demonstrates the advantage of our learning method.

\paragraph{Analysis of human annotations:} We check the distributions of human labels for DeepCom, Att-TC, and IR-TC to get insights on the problems these models suffer from. Table \ref{tab:score_dist} shows the results. Most of the bad comments from IR-TC are labeled as ``2'', meaning that although IR-TC can give attractive comments with rich and deep content, its comments are easy to diverge from the news articles and thus be judged as ``irrelevant''.  In terms of Att-TC, there are much more comments judged as ``1'' than the other two models, indicating that Att-TC often generates ill-formed sentences. This is because a news article and a comment are highly asymmetric in terms of both syntax and semantics, and thus the generation process cannot be simply modeled with an encoder-decoder structure. Bad cases from DeepCom concentrate on ``3'', reminding us that we need to further enrich the content of comments and improve their relevance in the future.

\begin{table}[t!]
\small
    \centering
    \resizebox{1\linewidth}{!}{
    \begin{tabular}{c r c c c c c}
        \toprule
                                  &  & 1 & 2 & 3 & 4 & 5\\
         \midrule
         \multirow{3}{*}{Tencent} & IR-TC  &  0.3\% & 55.2\% & 33.3\% & 9.50\% & 1.7\%\\
                                  & Att-TC & 19.3\% & 49.5\% & 18.5\% & 11.7\% & 1.0\%\\
                                  & DeepCom&  1.5\% &  3.3\% & 32.0\% & 61.9\% & 1.3\%\\
         \midrule
         \multirow{3}{*}{Yahoo}   & IR-TC  &  1.3\% & 53.0\% & 14.7\% & 20.3\% & 10.7\%\\
                                  & Att-TC & 20.3\% & 22.3\% & 22.6\% & 32.1\% & 2.7\%\\
                                  & DeepCom&  1.7\% & 17.6\% & 33.3\% & 39.4\% & 8.0\%\\
         \bottomrule
    \end{tabular}
    }
    \caption{Human score distributions}
    \label{tab:score_dist}
\end{table}

\paragraph{Case Study:} Finally, to further understand our model, we visualize the predicted salient spans and the generated comment using a test example from Tencent dataset in Table \ref{tab:showcase}. Due to space limitation, we truncate the body and only show three of the selected spans in the truncated body. The full article with the full set of spans and another test example from Yahoo! News dataset are shown in Appendix \ref{apd:case}. In spite of this, we can see that the model finds some interesting points after ``reading'' the article and synthesizes a comment along one of the spans (i.e. ``Chinese Paladin 3''). More interestingly, the model extends the content of the article in the comment with ``Luo Jin'',  who is Tiffany Tang's partner but is not talked about in the article. On the other hand, comments given by the baseline methods are either too general (Att-TC, the best generation baseline), or totally irrelevant with the article (IR-TC, the best retrieval baseline).  The example demonstrates that our model can generate relevant and informative comments through analyzing and understanding news articles.

\begin{CJK*}{UTF8}{gbsn} 
\begin{table}[!t]
\small
    \centering
    \begin{tabular}{p{7.3cm}}
    \toprule
    \textbf{Title:}\space 唐嫣为什么不演清宫剧？ (Why Tiffany Tang never plays a role in a drama set in Qing dynasty?)\\
    \midrule
    \textbf{Body (truncated):}\space众多的戏剧影视作品让唐嫣获得不少人气，众多活动的多变造型树立其在\mybox{大众心中百变小天后}的形象。...
    如果说\mybox{唐嫣最美的造型是《仙剑奇侠传三》中的紫萱}，\\ \mybox{那最丑的造型}应该就是这个了吧！... \\
    (... The numerous television series and movies Tiffany Tang acted in have made her very popular, and her varied modelling in many activities has set her an image of \mybox{"the queen of reinvention" in the hearts of the public}. ... If \mybox{her most beautiful role is Zi Xuan in ``Chinese Paladin 3''}, \mybox{then the most ugly one} should be this! ... )\\ 
    \midrule
    \textbf{DeepCom:}\space唐嫣罗晋的演技真的很好，特别喜欢她演\mybox[blue]{《仙剑奇侠传》}。我觉得这部剧挺好看。(Tiffany Tang and  Luo Jin are really good actors. I especially like her role in \mybox[blue]{``Chinese Paladin 3''}. I think the TV drama is worth watching.)\\
    \textbf{Att-TC:}\space我也是醉了 (I have nothing to comment.)\\
    \textbf{IR-TC:}\space星爷和谁开撕过嘛, 都是别人去撕星爷！(Stephen Chow never fights with others. It is others that fight with Stephen Chow!)\\
    \bottomrule
    \end{tabular}
    \caption{A Case from Tencent News dataset. The contents in the red box represent salient spans predicted by the reading network. The content in the blue box is a generated entity which is included in the salient spans.}
    \label{tab:showcase}
    \end{table}	

\end{CJK*}

%% file: conclusion.tex
\section{Conclusions}
We propose automatic news comment generation with a reading network and a generation network. 
Experimental results on two datasets indicate that the proposed model can significantly outperform baseline methods in terms of both automatic evaluation and human evaluation. On applications, we are motivated to extend the capabilities of a popular chatbot. We are aware of potential ethical issues with application of these methods to generate news commentary that is taken as human. We hope to stimulate discussion about best practices and controls on these methods around responsible uses of the technology.

\section*{Acknowledgment}
This work is supported in part by the National Natural Science Foundation of China (Grand Nos. U1636211, 61672081, 61370126), and the National Key R\&D Program of China (No. 2016QY04W0802).

%% file: appendix.tex
\appendix
\newpage
\onecolumn
\section{Yahoo! News Dataset}
\label{apd:data}
\begin{figure}[h!]
    \centering
    \includegraphics[width=3.5in]{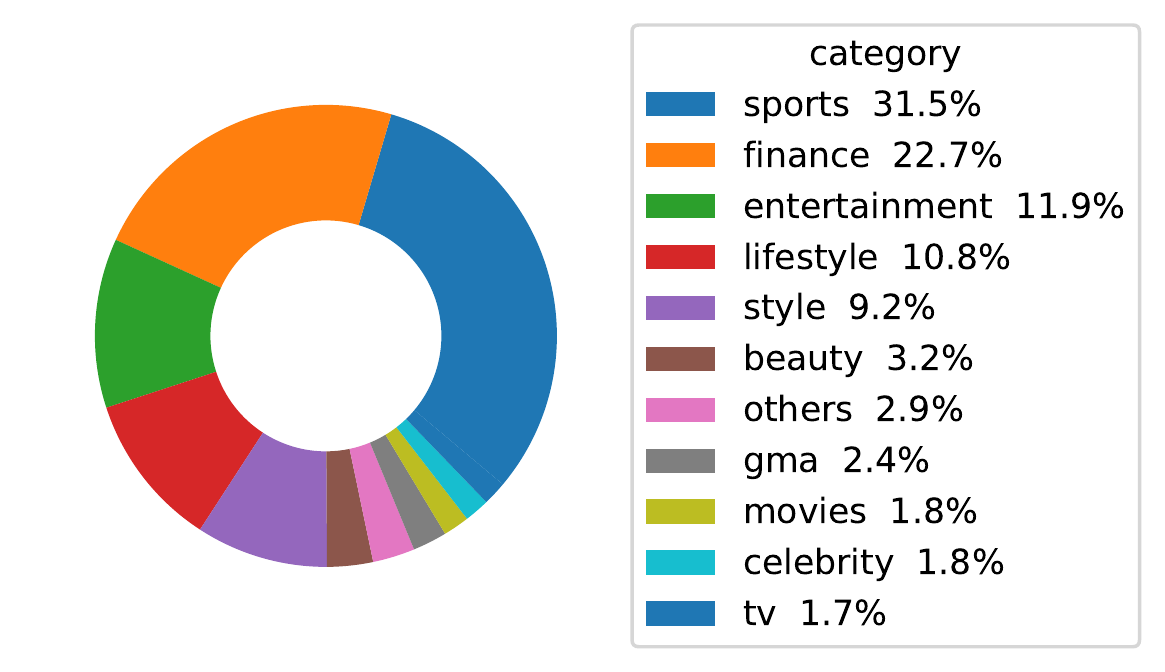}
    \caption{The category distribution of Yahoo! News dataset.}
    \label{fig:cate-dis}
\end{figure}

More information about the Yahoo dataset is shown in Table \ref{tab:yahoo-dataset}. The side information associated with news including: 
\begin{itemize}[itemsep=-0.5mm]
\item  \textbf{Paragraph}. After pre-processing, we retain the paragraph structure of news articles.
\item  \textbf{Category}. There are 31 news categories and the distribution is shown in Figure \ref{fig:cate-dis}.
\item  \textbf{Wiki-Entities}. The Wikipedia entities mentioned in the news articles are extracted.
\item  \textbf{Vote}. Each comment has upvote, downvote and abusevote information from news readers. 
\item  \textbf{Sentiment}. Each comment is annotated with POSITIVE, NEGATIVE or NEUTRAL by Yahoo!. 
\end{itemize}

\begin{table}[h!]
  \centering
  \resizebox{0.6\columnwidth}{!}{
  \begin{tabular}{c r c c c c}
  \toprule
                                 &             & Train    & Dev   & Test  \\
        \midrule
        \multirow{3}{*}{Tencent}  &      \# News & 191,502  & 5,000 & 1,610 \\
                                 &  Avg. \# Comments per News & 27       & 27    & 27    \\
                                 & Avg. \#Upvotes per Comment & 5.9      & 4.9   & 3.4   \\
        \midrule
        \multirow{5}{*}{Yahoo}  &          \# News & 152,355  & 5,000 & 3,160 \\
                                &     Avg. \# Comments per News & 20.6     & 20.5  & 20.5  \\
                                & Avg. \#Upvotes per Comment & 31.4     & 30.2  & 32.0  \\
                                & Avg. \#DownVotes per Comment & 4.8      & 4.8   & 4.9   \\
                                & Avg. \#AbuseVotes per Comment & 0.05     & 0.05  & 0.05  \\
        \bottomrule
  \end{tabular}
  }
  \caption{Statistics of the two datasets.  }
  \label{tab:yahoo-dataset}
\end{table}

\section{Case Study}
\label{apd:case}
We demonstrate the advantage of DeepCom over IR-TC and Att-TC with examples from the test sets of the two datasets. Table \ref{tab:tencent-case} and Table \ref{tab:yahoo-case} show the comments given by the three models. The content in the red box represents salient spans predicted by the reading network and that in the blue box is generated entities included in the salient spans. We can see that compared to the two baseline methods, comments from DeepCom are more relevant to the content of the news articles. Comments from IR-TC are rich with content, but are irrelevant to the news articles, while Att-TC is prone to generate generic comments. 
\begin{CJK*}{UTF8}{gbsn} 
  \begin{table*}[!ht]
  \small
      \centering
      \begin{tabular}{p{15cm}}
      \toprule
      \textbf{Title:}\space 唐嫣为什么不演清宫剧？ (Why Tiffany Tang never plays a role in a drama set in Qing dynasty?)\\
      \midrule
      \textbf{Body:}\space众多的戏剧影视作品让唐嫣博获得不少人气，众多活动的多变造型树立其在\mybox{大众心中“百变小天后”}的形象。商业活动高贵优雅、媒体活动清新脱俗，\mybox{唐嫣每走到一处总能获得交口称赞和镁光灯的钟爱}。\mybox{唐嫣身材高挑}、\mybox{笑容甜美}，\mybox{精致的}脸庞，时尚的装扮，有着“九头身美女”之称。如果说\mybox{唐嫣最美的造型是《仙剑奇侠传三》中的紫萱}，\mybox{那最丑的造型}应该就是这个了吧！无论是发型还是脸都无法直视啊！《仙剑奇侠传三》时候应该算是唐嫣的颜值巅峰期了吧！而穿上宫廷服装的唐嫣，真的\mybox{不好看，宛如一个智障少女……这一段清宫戏，是唐嫣在电视剧《乱世佳人》的造型，但}这部戏简直是\mybox{唐嫣不堪回首的过去啊！这个民国时期的造型也不好看！}终于明白为什么\mybox{唐嫣从来}\mybox{不演清宫剧了！如果不用假刘海}，就算脸在小，她的额头还是很大！对此，大家怎么看？\\
      (The numerous television series and movies Tiffany Tang acted in have made her very popular, and her varied modelling in many activities has set her an image of \mybox{"the queen of reinvention" in the hearts of the public}. Her elegant and unique nature showed in commercial events and media events \mybox{brings her a lot of praise and spotlight}.  \mybox{Tiffany Tang is tall}, \mybox{slim}, \mybox{has sweet smile and a beautiful} face, and is always dressed in fashion. Thus, she is called ``9-heads tall'' girl by her fans. If \mybox{her most beautiful role is Zi Xuan in ``Chinese Paladin 3''}, \mybox{then the most ugly one} should be this! Both hair-style and makeup look terrible. She is most alluring at the time of ``Chinese Paladin 3''. When she dressed up with a costume in Qing Dynasty, however, she \mybox{looks really bad, like an idiot. This is how Tiffany Tang looks like  in a TV series ``A Beauty in Troubled Times'', but} this might be her forgettable moment. Her costume in Republic of China looks terrible too. I finally know why \mybox{Tiffany Tang never plays a role in a drama set in Qing dynasty. If she does not use fake bangs}, even her face is small, her forehead looks too big! What do you think?)\\
      \midrule
      \textbf{DeepCom:}\space唐嫣罗晋的演技真的很好，特别喜欢她演\mybox[blue]{《仙剑奇侠传》}。我觉得这部剧挺好看。(Tiffany Tang and  Luo Jin are really good actors. I especially like her role in \mybox[blue]{Chinese Paladin 3}. I think the TV drama is worth watching.)\\
      \textbf{Att-TC:}\space我也是醉了 (I have nothing to comment.)\\
      \textbf{IR-TC:}\space星爷和谁开撕过嘛, 都是别人去撕星爷！（Stephen Chow never fights with others. It is others that fight with Stephen Chow!）\\
      \bottomrule
      \end{tabular}
      \caption{A case from Tencent News dataset. }
      \label{tab:tencent-case}
      \end{table*}	
  \end{CJK*}

\begin{table*}[!ht]
  \small
      \centering
      \begin{tabular}{p{15cm}}
      \toprule
      \textbf{Title:}\space NBA notebook : Rockets targeting Anthony after losing Mbah a Moute\\
      \midrule
      \textbf{Body:}\space The Houston Rockets are now determined to \mybox{sign forward Carmelo Anthony after forward Luc Mbah a Moute} \mybox{joined the Los Angeles Clippers} on a one-year, \$4.3 million deal on Monday, according to an ESPN report. Anthony is currently a member of the \mybox{Oklahoma City Thunder}, but the two sides are reportedly working on parting ways, whether through a trade, a buyout or waiving via the stretch provision. Anthony is likely to become a free agent even if he is traded, as his new team would likely waive him. Multiple reports on Sunday said \mybox{rockets} guard Chris Paul wants Anthony, a \mybox{good friend of his}, to join the rockets, \mybox{while Anthony is also believed to have interest} \mybox{in joining Lebron James with the Los Angeles Lakers.} The \mybox{Miami Heat are} also reportedly interested in adding Anthony. \mybox{Mbah a Moute spent the 2015-16 and 2016-17 seasons with the Clippers} before joining the Rockets last season. \mybox{The 31-year-old averaged 7.5 points , 3.0 rebounds} and 1.2 steals in 25.6 minutes per game across 61 games (15 starts) in Houston.  -- The Cleveland cavaliers are looking to deal 37-year-old guard Kyle Korver and transition to a younger lineup, according to Terry Pluto of the Cleveland Plain Dealer. \ldots Korver's contract has \$ 15.1 million remaining over the next two seasons, although only \$ 3.4 million of his 2019-20 salary is guaranteed. He could prove to be a \mybox{good option for a team looking for better perimeter shooting to compete for a league title}. \ldots -- It 's unclear whether Joakim Noah will remain with the New York Knicks moving forward, but the center said he hopes to say in the big apple, in a video published by TMZ. ``\mybox{I love New York," Noah said}, ``I don't know what's going to happen , but Coach Fiz is cool, man." \ldots \\ 
      \midrule
      \textbf{DeepCom:}\space the \mybox[blue]{rockets} are going to have a lot of fun in this series .\\
      \textbf{Att-TC:}\space i think he is going to be a contender\\
      \textbf{IR-TC:}\space the kings have a great future if cousins is n't the team 's leader . but if continues to be propped up as the guy , then the kings will only have a good future along with the bottom half of the west . cousins just is n't enough of a leadership guy who will go through 3 more coaches before he finds a decent match not just for himself , but the team .\\
      \bottomrule 
      \end{tabular}
      \caption{A case from Yahoo! News dataset.}
      \label{tab:yahoo-case}
      \end{table*}	

%% file: main.bbl
\begin{thebibliography}{31}
\expandafter\ifx\csname natexlab\endcsname\relax\def\natexlab#1{#1}\fi

\bibitem[{van Aken et~al.(2018)van Aken, Risch, Krestel, and
  L{\"o}ser}]{van2018challenges}
Betty van Aken, Julian Risch, Ralf Krestel, and Alexander L{\"o}ser. 2018.
\newblock Challenges for toxic comment classification: An in-depth error
  analysis.
\newblock \emph{arXiv preprint arXiv:1809.07572}.

\bibitem[{Bahdanau et~al.(2015)Bahdanau, Cho, and Bengio}]{bahdanau2014neural}
Dzmitry Bahdanau, Kyunghyun Cho, and Yoshua Bengio. 2015.
\newblock Neural machine translation by jointly learning to align and
  translate.
\newblock In \emph{International Conference on Learning Representations}.

\bibitem[{Banerjee and Lavie(2005)}]{banerjee2005meteor}
Satanjeev Banerjee and Alon Lavie. 2005.
\newblock \href {http://aclweb.org/anthology/W05-0909} {Meteor: An automatic
  metric for mt evaluation with improved correlation with human judgments}.
\newblock In \emph{Proceedings of the ACL Workshop on Intrinsic and Extrinsic
  Evaluation Measures for Machine Translation and/or Summarization}, pages
  65--72. Association for Computational Linguistics.

\bibitem[{Chen and Bordes(2017)}]{chen2017reading}
Danqi Chen and Antoine Bordes. 2017.
\newblock \href {https://doi.org/10.18653/v1/P17-1171} {Reading wikipedia to
  answer open-domain questions}.
\newblock In \emph{Proceedings of the 55th Annual Meeting of the Association
  for Computational Linguistics (Volume 1: Long Papers)}, pages 1870--1879.

\bibitem[{Cho et~al.(2014)Cho, van Merri{\"e}nboer, Bahdanau, and
  Bengio}]{cho2014properties}
Kyunghyun Cho, Bart van Merri{\"e}nboer, Dzmitry Bahdanau, and Yoshua Bengio.
  2014.
\newblock \href {https://doi.org/10.3115/v1/W14-4012} {On the properties of
  neural machine translation: Encoder--decoder approaches}.
\newblock \emph{Proceedings of SSST-8, Eighth Workshop on Syntax, Semantics and
  Structure in Statistical Translation}, pages 103--111.

\bibitem[{Dong et~al.(2017)Dong, Huang, Wei, Lapata, Zhou, and
  Xu}]{dong2017learning}
Li~Dong, Shaohan Huang, Furu Wei, Mirella Lapata, Ming Zhou, and Ke~Xu. 2017.
\newblock \href {http://aclweb.org/anthology/E17-1059} {Learning to generate
  product reviews from attributes}.
\newblock In \emph{Proceedings of the 15th Conference of the European Chapter
  of the Association for Computational Linguistics: Volume 1, Long Papers},
  pages 623--632.

\bibitem[{Duchi et~al.(2011)Duchi, Hazan, and Singer}]{duchi2011adaptive}
John Duchi, Elad Hazan, and Yoram Singer. 2011.
\newblock \href {http://dl.acm.org/citation.cfm?id=1953048.2021068} {Adaptive
  subgradient methods for online learning and stochastic optimization}.
\newblock \emph{J. Mach. Learn. Res.}, 12:2121--2159.

\bibitem[{Gatt and Krahmer(2018)}]{gatt2018survey}
Albert Gatt and Emiel Krahmer. 2018.
\newblock Survey of the state of the art in natural language generation: Core
  tasks, applications and evaluation.
\newblock \emph{Journal of Artificial Intelligence Research}, 61:65--170.

\bibitem[{Lin(2004)}]{lin2004rouge}
Chin-Yew Lin. 2004.
\newblock \href {http://aclweb.org/anthology/W04-1013} {Rouge: A package for
  automatic evaluation of summaries}.
\newblock \emph{Text Summarization Branches Out}.

\bibitem[{Manning et~al.(2014)Manning, Surdeanu, Bauer, Finkel, Bethard, and
  McClosky}]{manning-EtAl:2014:P14-5}
Christopher~D. Manning, Mihai Surdeanu, John Bauer, Jenny Finkel, Steven~J.
  Bethard, and David McClosky. 2014.
\newblock \href {http://www.aclweb.org/anthology/P/P14/P14-5010} {The
  {Stanford} {CoreNLP} natural language processing toolkit}.
\newblock In \emph{Association for Computational Linguistics (ACL) System
  Demonstrations}, pages 55--60.

\bibitem[{Mnih and Gregor(2014)}]{mnih2014neural}
Andriy Mnih and Karol Gregor. 2014.
\newblock \href {http://dl.acm.org/citation.cfm?id=3044805.3045092} {Neural
  variational inference and learning in belief networks}.
\newblock In \emph{Proceedings of the 31st International Conference on
  International Conference on Machine Learning - Volume 32}, ICML'14, pages
  II--1791--II--1799. JMLR.org.

\bibitem[{Nallapati et~al.(2016)Nallapati, Zhou, dos Santos, glar
  Gul{\c{c}}ehre, and Xiang}]{nallapati2016abstractive}
Ramesh Nallapati, Bowen Zhou, Cicero dos Santos, {\c{C}}a~glar Gul{\c{c}}ehre,
  and Bing Xiang. 2016.
\newblock \href {https://doi.org/10.18653/v1/K16-1028} {Abstractive text
  summarization using sequence-to-sequence rnns and beyond}.
\newblock pages 280--290.

\bibitem[{Nguyen et~al.(2016)Nguyen, Rosenberg, Gao, and Deng}]{nguyen2016ms}
Tri Nguyen, Mir Rosenberg, Jianfeng Gao, and Li~Deng. 2016.
\newblock Ms marco: A human generated machine reading comprehension dataset.
\newblock \emph{arXiv preprint arXiv:1611.09268}.

\bibitem[{Papineni et~al.(2002)Papineni, Roukos, Ward, and
  Zhu}]{papineni2002bleu}
Kishore Papineni, Salim Roukos, Todd Ward, and Wei-Jing Zhu. 2002.
\newblock \href {http://aclweb.org/anthology/P02-1040} {Bleu: a method for
  automatic evaluation of machine translation}.
\newblock In \emph{Proceedings of the 40th Annual Meeting of the Association
  for Computational Linguistics}, pages 311--318.

\bibitem[{Qin et~al.(2018)Qin, Liu, Bi, Wang, Liu, Hu, Zhao, and
  Shi}]{qin2018automatic}
Lianhui Qin, Lemao Liu, Wei Bi, Yan Wang, Xiaojiang Liu, Zhiting Hu, Hai Zhao,
  and Shuming Shi. 2018.
\newblock Automatic article commenting: the task and dataset.
\newblock \emph{arXiv preprint arXiv:1805.03668}.

\bibitem[{Rajpurkar et~al.(2018)Rajpurkar, Jia, and Liang}]{rajpurkar2018know}
Pranav Rajpurkar, Robin Jia, and Percy Liang. 2018.
\newblock Know what you don't know: Unanswerable questions for squad.
\newblock \emph{arXiv preprint arXiv:1806.03822}.

\bibitem[{Rajpurkar et~al.(2016)Rajpurkar, Zhang, Lopyrev, and
  Liang}]{rajpurkar2016squad}
Pranav Rajpurkar, Jian Zhang, Konstantin Lopyrev, and Percy Liang. 2016.
\newblock \href {https://doi.org/10.18653/v1/D16-1264} {{SQuAD}: 100,000+
  questions for machine comprehension of text}.
\newblock In \emph{Proceedings of the 2016 Conference on Empirical Methods in
  Natural Language Processing}, pages 2383--2392.

\bibitem[{Rush et~al.(2015)Rush, Chopra, and Weston}]{rush2015neural}
Alexander~M Rush, Chopra, and Jason Weston. 2015.
\newblock \href {https://doi.org/10.18653/v1/D15-1044} {A neural attention
  model for abstractive sentence summarization}.
\newblock In \emph{Proceedings of the 2015 Conference on Empirical Methods in
  Natural Language Processing}, pages 379--389.

\bibitem[{See et~al.(2017)See, Liu, and Manning}]{see2017get}
Abigail See, Peter~J Liu, and Christopher~D Manning. 2017.
\newblock \href {https://doi.org/10.18653/v1/P17-1099} {Get to the point:
  Summarization with pointer-generator networks}.
\newblock In \emph{Proceedings of the 55th Annual Meeting of the Association
  for Computational Linguistics (Volume 1: Long Papers)}, pages 1073--1083.

\bibitem[{Seo et~al.(2016)Seo, Kembhavi, Farhadi, and
  Hajishirzi}]{seo2016bidirectional}
Minjoon Seo, Aniruddha Kembhavi, Ali Farhadi, and Hannaneh Hajishirzi. 2016.
\newblock Bidirectional attention flow for machine comprehension.
\newblock \emph{arXiv preprint arXiv:1611.01603}.

\bibitem[{Shum et~al.(2018)Shum, He, and Li}]{shum2018eliza}
Heung{-}Yeung Shum, Xiaodong He, and Di~Li. 2018.
\newblock \href {https://doi.org/10.1631/FITEE.1700826} {From eliza to xiaoice:
  Challenges and opportunities with social chatbots}.
\newblock \emph{Frontiers of Information Technology {\&} Electronic
  Engineering}, 19(1):10--26.

\bibitem[{Sutskever et~al.(2014)Sutskever, Vinyals, and
  Le}]{sutskever2014sequence}
Ilya Sutskever, Oriol Vinyals, and Quoc~V. Le. 2014.
\newblock \href {http://dl.acm.org/citation.cfm?id=2969033.2969173} {Sequence
  to sequence learning with neural networks}.
\newblock In \emph{Proceedings of the 27th International Conference on Neural
  Information Processing Systems - Volume 2}, NIPS'14, pages 3104--3112,
  Cambridge, MA, USA. MIT Press.

\bibitem[{Tan et~al.(2018)Tan, Wei, Yang, Du, Lv, and Zhou}]{tan2017s}
Chuanqi Tan, Furu Wei, Nan Yang, Bowen Du, Weifeng Lv, and Ming Zhou. 2018.
\newblock S-net: From answer extraction to answer generation for machine
  reading comprehension.
\newblock In \emph{AAAI'18}.

\bibitem[{Tang et~al.(2016)Tang, Yang, Carton, Zhang, and
  Mei}]{tang2016context}
Jian Tang, Yifan Yang, Sam Carton, Ming Zhang, and Qiaozhu Mei. 2016.
\newblock Context-aware natural language generation with recurrent neural
  networks.
\newblock \emph{arXiv preprint arXiv:1611.09900}.

\bibitem[{Vaswani et~al.(2017)Vaswani, Shazeer, Parmar, Uszkoreit, Jones,
  Gomez, Kaiser, and Polosukhin}]{vaswani2017attention}
Ashish Vaswani, Noam Shazeer, Niki Parmar, Jakob Uszkoreit, Llion Jones,
  Aidan~N Gomez, \L~ukasz Kaiser, and Illia Polosukhin. 2017.
\newblock \href
  {http://papers.nips.cc/paper/7181-attention-is-all-you-need.pdf} {Attention
  is all you need}.
\newblock In I.~Guyon, U.~V. Luxburg, S.~Bengio, H.~Wallach, R.~Fergus,
  S.~Vishwanathan, and R.~Garnett, editors, \emph{Advances in Neural
  Information Processing Systems 30}, pages 5998--6008. Curran Associates, Inc.

\bibitem[{Vedantam et~al.(2015)Vedantam, Lawrence~Zitnick, and
  Parikh}]{vedantam2015cider}
Ramakrishna Vedantam, C~Lawrence~Zitnick, and Devi Parikh. 2015.
\newblock Cider: Consensus-based image description evaluation.
\newblock In \emph{The IEEE Conference on Computer Vision and Pattern
  Recognition (CVPR)}, pages 4566--4575.

\bibitem[{Vinyals et~al.(2015)Vinyals, Fortunato, and
  Jaitly}]{vinyals2015pointer}
Oriol Vinyals, Meire Fortunato, and Navdeep Jaitly. 2015.
\newblock \href {http://papers.nips.cc/paper/5866-pointer-networks.pdf}
  {Pointer networks}.
\newblock In C.~Cortes, N.~D. Lawrence, D.~D. Lee, M.~Sugiyama, and R.~Garnett,
  editors, \emph{Advances in Neural Information Processing Systems 28}, pages
  2692--2700. Curran Associates, Inc.

\bibitem[{Wang et~al.(2017)Wang, Yang, Wei, Chang, and Zhou}]{wang2017gated}
Wenhui Wang, Nan Yang, Furu Wei, Baobao Chang, and Ming Zhou. 2017.
\newblock \href {https://doi.org/10.18653/v1/P17-1018} {Gated self-matching
  networks for reading comprehension and question answering}.
\newblock In \emph{Proceedings of the 55th Annual Meeting of the Association
  for Computational Linguistics (Volume 1: Long Papers)}, volume~1, pages
  189--198.

\bibitem[{Xiong et~al.(2016)Xiong, Zhong, and Socher}]{xiong2016dynamic}
Caiming Xiong, Victor Zhong, and Richard Socher. 2016.
\newblock Dynamic coattention networks for question answering.
\newblock \emph{arXiv preprint arXiv:1611.01604}.

\bibitem[{Yu et~al.(2018)Yu, Dohan, Chen, Norouzi, and Le}]{yu2018qanet}
Adams~Wei Yu, Rui Dohan, Kai Chen, Mohammad Norouzi, and Quoc~V Le. 2018.
\newblock \href {https://openreview.net/forum?id=B14TlG-RW} {Qanet: Combining
  local convolution with global self-attention for reading comprehension}.
\newblock In \emph{International Conference on Learning Representations}.

\bibitem[{Zheng et~al.(2018)Zheng, Wang, Chen, and
  Sangaiah}]{zheng2018automatic}
Hai-Tao Zheng, Wei Wang, Wang Chen, and Arun~Kumar Sangaiah. 2018.
\newblock \href {https://doi.org/10.1109/ACCESS.2017.2774839} {Automatic
  generation of news comments based on gated attention neural networks}.
\newblock \emph{IEEE Access}, 6:702--710.

\end{thebibliography}
